\documentclass[sigconf]{acmart}

\AtBeginDocument{%
  \providecommand\BibTeX{{%
    \normalfont B\kern-0.5em{\scshape i\kern-0.25em b}\kern-0.8em\TeX}}}

\setcopyright{acmcopyright}
\copyrightyear{2022}
\acmYear{2022}
\usepackage{graphicx}
\usepackage{multirow}
\usepackage{float}

\acmConference[ICCA '22,]{2nd International Conference on Computing Advancements,}{March 10-12, 2022}{Dhaka, Bangladesh}

\copyrightyear{2022}
\acmYear{2022}
\setcopyright{acmcopyright}\acmConference[ICCA 2022]{2nd International Conference on Computing Advancements}{March 10--12, 2022}{Dhaka, Bangladesh}
\acmBooktitle{2nd International Conference on Computing Advancements (ICCA 2022), March 10--12, 2022, Dhaka, Bangladesh}
\acmPrice{15.00}
\acmDOI{10.1145/3542954.3543017}
\acmISBN{978-1-4503-9734-6/22/03}

\begin{document}

\title{Predicting skull fractures via CNN with classification algorithms}

%

\author{Md Moniruzzaman Emon}
\affiliation{%
	\institution{Department
 of Computer Science and Engineering, Shahjalal
University of Science and Technology}
	\city{Sylhet}
	\country{Bangladesh}}
\email{moniruzzaman5673@gmail.com}

\author{ Tareque Rahman Ornob}
\affiliation{%
  \institution{Department
 of Computer Science and Engineering, Shahjalal
University of Science and Technology}
  \city{Sylhet}
  \country{Bangladesh}}
\email{ornob011@gmail.com}

\author{Moqsadur Rahman}
\affiliation{%
  \institution{Department of Computer Science and Engineering, Shahjalal University of Science and Technology}
  \city{Sylhet}
  \country{Bangladesh}}
\email{moqsad-cse@sust.edu}

%
\renewcommand{\shortauthors}{M. M. Emon et al.}

%
\begin{abstract}
Computer Tomography (CT) images have become quite important to diagnose diseases. CT scan slice contains a vast amount of data that may not be properly examined with the requisite precision and speed using normal visual inspection. A computer-assisted skull fracture classification expert system is needed to assist physicians. Convolutional Neural Networks (CNNs) are the most extensively used deep learning models for image categorization since most often time they outperform other models in terms of accuracy and results. The CNN models were then developed and tested, and several convolutional neural network (CNN) architectures were compared. ResNet50, which was used for feature extraction combined with a gradient boosted decision tree machine learning algorithm to act as a classifier for the categorization of skull fractures from brain CT scans into three fracture categories, had the best overall F1-score of 96\%, Hamming Score of 95\%, Balanced accuracy Score of 94\% \& ROC AUC curve of 96\% for the classification of skull fractures.

\end{abstract}





\maketitle

\section{Introduction}
Because a significant hit or blow to the head might result in a skull fracture as well as a brain injury, it's critical to figure out what kind of brain injury the skull fracture might cause as soon as possible so that the patient can get the care he or she needs. If the fracture occurs over a major blood vessel, significant bleeding may occur inside the brain, so head injury patients with skull fractures have far more intracranial hematomas than those without fractures.\cite{zaki2008automated,liu2008hemorrhage} Classification is a method of identifying the lesion caused by skull fractures\cite{national2007head}. Computed tomography (CT) has become the primary diagnostic tool for suspected skull or brain injuries.CT images and radiology reports, in general, provide more information to a physician, allowing them to make an informed decision. In the typical diagnosis approach, the radiologist examines the image and notes the results, and the physician then chooses a treatment based on the diagnosis. All of this takes a lot of time. 
A radiologist uses a CT scan to assess whether there is a skull fracture and which category it belongs to. On CT scans, however, the skull fracture has the following characteristics: fractures typically appear as narrow slits, fractures can be seen in a variety of locations and lengths, and a large proportion of fractures are microscopic.
All of these factors might make manual diagnosis and grading of skull fractures time-consuming and challenging. As a result, it's vital to propose a reliable automated skull fracture categorization and detection system. Skull fractures might be automatically detected and classified, which could aid in the diagnosis of other anomalies in CT scan brain imaging. Furthermore, many hospitals around the world are understaffed, resulting in delays in evaluating CT scan images.
In the CQ-500 dataset \cite{CQ-500}, their three radiologists couldn't agree on whether a patient's skull was fractured or normal in many situations, much alone classify the fracture as a calvarial fracture or another fracture deterministically. Automatic classification could assist clinicians in a short-staffed hospital in identifying the most critical patients and prioritizing their treatment.
Information contained in the  Computed  Tomography  (CT)  images is very important for assessing the severity and prognosis of the Classification of skull fracture. Each skull CT scan includes a large number of slices. 
To give clinicians with suggestions and predictions on diagnostic judgments and treatment planning, highly efficient and automated computational approaches are urgently needed to process and analyze all accessible medical data.
These facts provide the motivation for this work.
In addition to detecting fractured or normal cases from CT scan pictures, we provide a model for automated detection of common three skull fractures. These advantages could provide a good platform for retrieving content-based medical images for medical instruction or diagnosis
\section{Related Works}
Some approaches for identifying skull fractures have already been proposed.  Shao et al.\cite{shao2003automatic} have focused their efforts on CT brain segmentation for automated skull fracture diagnosis. They proposed utilizing a region-growing approach to segment the brain image and then using the entropy feature to construct guidelines for identifying skull fractures. This strategy has received a lot of positive feedback. Its complexity and performance, on the other hand, can be simplified and improved computationally. Zaki et al.\cite{za20ki09new} used Sobel edge detection technique for the diagnosis of skull fractures.
Despite the fact that the Sobel edge detection approach is superior in a variety of ways, it does yield some misshaping lines in some cases. It's worth mentioning that this method is incapable of handling large features. Prewitt is a straightforward approach to detect the boundaries based on gradient magnitude. However, when the amplitude of the gradient decreases, the accuracy will almost certainly decrease. Abubacker et al.\cite{abubacker2013approach} used histogram-based thresholding and nearby pixel connection search, presented a simple and fast automatic solution in Digital Imaging and Communications in Medicine (DlCOM) to extract the skull bone and diagnose the fracture. The experimental results for this approach are consistent, with a high detection rate. Chilamkurthy et al.\cite{chilamkurthy2018deep} have developed a deep learning approach for detecting cerebral bleeding and its kinds (intraparenchymal, intraventricular, subdural, extradural, and subarachnoid), as well as calvarial fractures, midline shift, and mass effect. They demonstrated that deep learning systems were capable of doing this task with excellent precision. 
Emon et al.\cite{Emon_2022} developed a model called SkullNetV1 to classify seven skull fractures. However, SkullNetV1 lacks score in some classes. Kuang et al.\cite{kuang2020skull} have proposed a way for more precisely detecting skull fractures in a short period of time. Skull R-CNN is the name of the proposed approach. Skull R-CNN has fewer false positives than earlier research on skull fracture diagnosis while keeping good sensitivity. Yamada et al.\cite{yamada2016preliminary} developed a unique approach for automatically detecting linear skull fractures on head CT images by recognizing crack lines They used two types of phantoms to do a rudimentary examination. A crack line with a width of 0.35 mm was found in their experiment with a digital phantom.  

All of the methods outlined above totally focused on the skull's local properties. As far as we know, there are just two methods for detecting linear skull fractures automatically. Only one method exists to automate the detection of skull fractures but no one has automated the classification of Linear and Depressed skull fractures with very high accuracy.

\section{DataSet}
With respective permission, we collected 142 patients' head CT scan from Medinova Medical Services Ltd.\cite{medinova} and Ibn Sina Hospital Sylhet Limited.\cite{ibnsina} Every CT scan image was in DICOM format, with 512x512 pixels in slice thicknesses of 0.75 mm, 1.0 mm, and 5.0 mm. The amount of 1.0 mm slices was 25678, those were used to construct the dataset. The CT images of all 142 patients were evaluated and annotated by expert radiologists. 
First, the radiologists divided the dataset into the fracture and normal cases (8628 CT scan slices), then the fractured instances were further categorized of into two categories i.e. Linear Fracture (4578 CT scan slices) and Depressed Fracture (8492 CT scan slices). The paired data (15189 CT scan slices and radiology reports of 107 patients) were used as training sets. The remaining 35 CT scans (6509 CT scan slices) were used as test sets while the associated radiology reports were used to evaluate the predictions. The data distribution between classes was moderately imbalanced.
\begin{figure}[h]
  \centering
  
  \includegraphics[width=\linewidth]{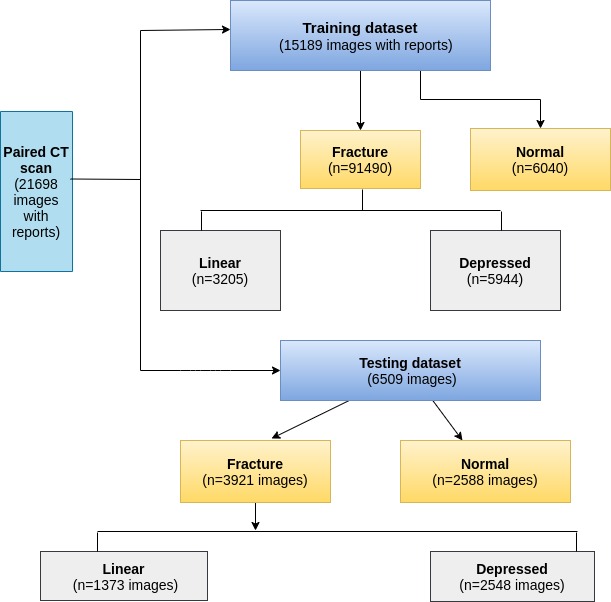}
 
  \Description{Data characteristics: Number of CT images and
radiology reports for training and testing the system.}
 \caption{Data characteristics. Number of CT images and
radiology reports for training and testing the system.}
  \label{Fig.1}
\end{figure}

\section{Methodology}
\subsection{Data Pre-processing}
\begin{figure}[h]
  \centering
 
  \includegraphics[width=\linewidth]{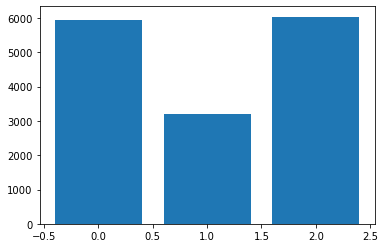}
 
  \Description{Data Distribution across the classes before balancing}
    \caption{Data Distribution across the classes before balancing}
  \label{Fig.2}
\end{figure}

1 mm DICOM sequence from each DICOM series was extracted automatically. Best DICOM slices were annotated by the respective radiologists. During the preparation of images, several processes were taken.
Transforming to HU, Removing Noises, Tilt Correction, Cropping Images, and Padding are the steps. Model accuracy improved significantly when these preprocessing techniques were applied to data. We converted our DICOM picture data to Hounsfield Unit form after loading it. Using Rescale Intercept and Rescale Slope headings, we were able to retrieve the HU. Removing noises is critical since the data is better after implementation, allowing us to view it more clearly. Tilt correction is the proposed alignment of the brain picture. When brain CT pictures are tilted, it might cause misalignment in medical applications. It's significant since the model can see all of the data via the same alignment when it's being trained. Cropping an image is required to center the brain image and remove superfluous sections of the image. Additionally, within the overall image, various brain images may be positioned in different locations. We ensured that almost all of the images are in the same area inside the general image by cropping it and adding pads. Each patient’s CT scan picture slices were given a unique ID in a CSV file along with the slice class i.e. Linear Fracture, Depressed Fracture, or normal cases in another column. LabelEncoder was used to transform the string format of the patients’ condition class into a one-row matrix i.e. 0, 1, 2. The encoded matrix was then converted into a hot one encoded matrix by LabelBinarizer. Every CT scan image was stored in a NumPy array. 
Data was highly imbalanced across the classes as shown in the \hyperref[Fig.2]{Figure 2:}\\
\textbf{Class=0 (Depressed Fracture), n=5944 (39.134\%)}\\
\textbf{Class=1 (Linear Fracture), n=3205 (21.101\%)}\\
\textbf{Class=2 (Not Fractured), n=6040 (39.766\%)}\\

To tackle this problem, two resampling techniques were implemented, Random oversampling and Random undersampling. Random oversampling involves randomly duplicating examples in the minority class, whereas random undersampling involves randomly deleting examples from the majority class. A pipeline was defined that first oversamples the minority class and under samples the majority class.

\begin{figure}[h]
  \centering
   
  \includegraphics[width=\linewidth]{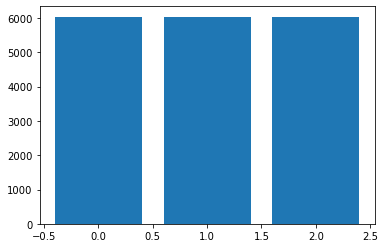}
 
  \Description{Data Distribution across the classes after balancing
}
\caption{Data Distribution across the classes after balancing
}
  \label{Fig.3}
\end{figure}

After applying this technique, the dataset became balanced as shown in the \hyperref[Fig.3]{Figure 3:}

\begin{figure}[h]
  \centering
   
  \includegraphics[width=\linewidth]{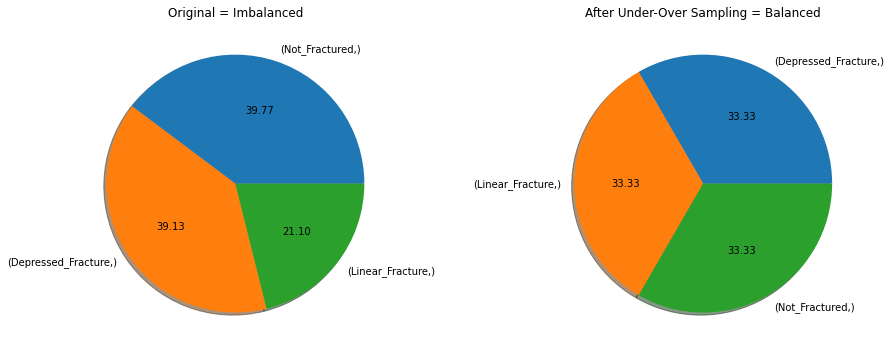}
 
  \Description{Data Distribution across the classes before & after balancing
}
\caption{Data Distribution across the classes before \& after balancing}
  \label{Fig.4}
\end{figure}

\textbf{Class=0 (Depressed Fracture), n=6040 (33.333\%)}\\
\textbf{Class=1 (Linear Fracture), n=6040 (33.333\%)}\\
\textbf{Class=2 (Not Fractured), n=6040 (33.333\%)}\\
A pie chart is given in \hyperref[Fig.4]{Figure 4} to show the full transformation.

\subsection{Data Visualization}
The estimated (Gaussian) noise standard deviation is 0.02 in our dataset. So, our data is almost noise-free, no other denoising technique is necessary to further denoise the data. To visualize our statement clearly, various noise filtering algorithm was performed on the dataset.
\begin{figure}[h]
  \centering
   
  \includegraphics[width=\linewidth]{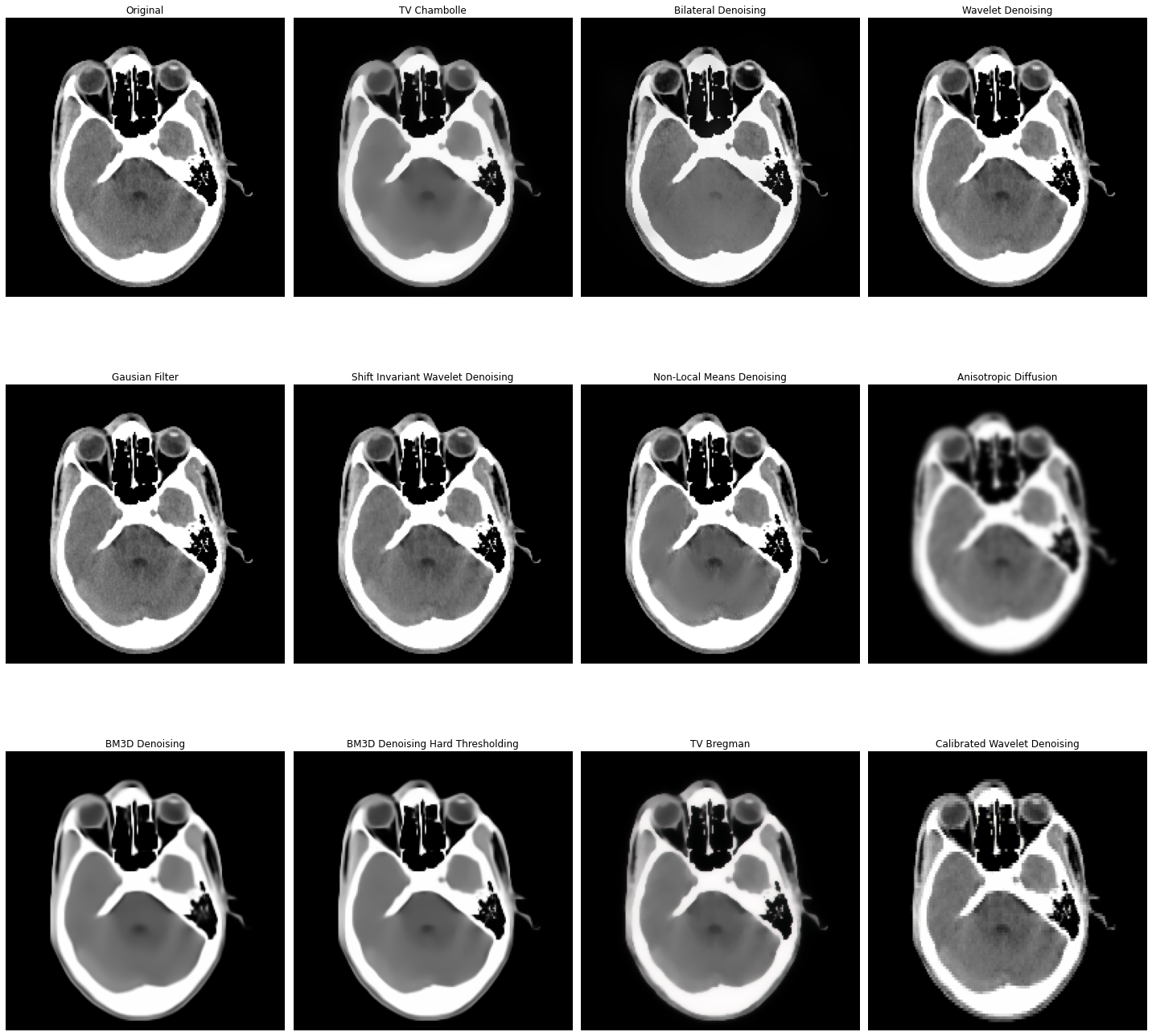}
 
  \Description{Visualization of an original random 2d slice after performing denoising

}
\caption{Visualization of an original random 2d slice after performing denoising}
  \label{Fig.5}
\end{figure}

We can see from \hyperref[Fig.5]{Figure 5} our original data is much clearer than after performing denoising filters.
\\
\\
\subsection{Implementation and Graph}
From different model-based approaches, we implemented:\\
\textbf{Convolutional Neural Network (Transfer Learning Based)}\\
\textbf{Bagging algorithm}\\
\textbf{Decision Tree}\\
\textbf{Support Vector Machine}\\

Convolutional Neural Network:\\
Xception, InceptionV3, ResNet50, and InceptionResNetV2 was our preferred CNN model due to their state-of-the-art architecture. \\
Every model CNN was imported from TensorFlow and was fine-tuned by:
\begin{itemize}
  \item Loading weights from available pre-trained models, included with Keras library. 
  \item Stacking another network for training on top of any layers 
  \item Inserting a layer in the middle of other layers
  \item Freezing multiple layers
  \item Removing multiple layers and inserting a new one in the middle
\end{itemize}
Learning Rate was kept low for Nadam, Adam, and SGD, momentum was used to avoid local minimum. Categorical Crossentropy was used as the loss function, the activation function was relu and the activation function in the output layer was softmax. F1-score was chosen as the metric. The dense layer was composed of 128 and 3 units, with l2 kernel regulizer for InceptionV3.
\\
\\
\textbf{Training vs Validation Loss Graphs:}\\

Xception:\\
\begin{figure}[h]
  \centering
  
  \includegraphics[width=\linewidth]{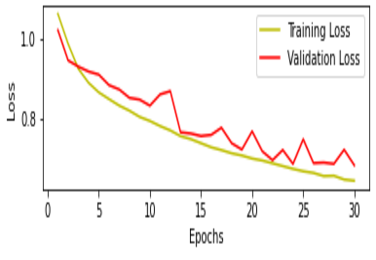}
 
  \Description{Xception}
   \caption{Cross Entropy loss of Xception}
  \label{Fig.6}
\end{figure}
Xception performed reasonably well until the 30 epoch. We can see that the validation curve is fluctuating because the validation dataset is quite small compared to the training dataset.\\

\begin{table}[H]
\caption{Score of Xception}
\begin{tabular}{|l|l|l|}
\hline
\multicolumn{1}{|c|}{Model} & \multicolumn{1}{c|}{Metric} & \multicolumn{1}{c|}{Score} \\ \hline
\multirow{7}{*}{Xception}   & F1 score (micro avg)        & 0.37                       \\ \cline{2-3} 
 & Hamming Score           & 0.36  \\ \cline{2-3} 
 & Hamming Loss            & 0.42  \\ \cline{2-3} 
 & Balanced Accuracy Score & 0.39  \\ \cline{2-3} 
 & ROC AUC                 & 0.55  \\ \cline{2-3} 
 & Kappa Score             & 0.105 \\ \cline{2-3} 
 & Log Loss                & 21.87 \\ \hline
\end{tabular}
\label{Tab.1}
\end{table}









InceptionV3:\\
The same result as Xception, but validation loss is less fluctuating.\\
\begin{figure}[h]
  \centering
 
  \includegraphics[width=\linewidth]{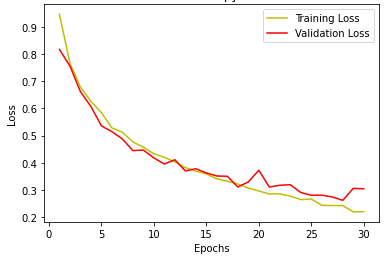}
 
  \Description{Inceptionv3
}
  \caption{Cross Entropy loss of InceptionV3
}
  \label{Fig.8}
\end{figure}

\begin{table}[H]
\caption{Score of InceptionV3}
\begin{tabular}{|l|l|l|}
\hline
\multicolumn{1}{|c|}{Model} & \multicolumn{1}{c|}{Metric} & \multicolumn{1}{c|}{Score} \\ \hline
\multirow{7}{*}{InceptionV3}   & F1 score (micro avg)        & 0.37                       \\ \cline{2-3} 
 & Hamming Score           & 0.37  \\ \cline{2-3} 
 & Hamming Loss            & 0.42  \\ \cline{2-3} 
 & Balanced Accuracy Score & 0.38  \\ \cline{2-3} 
 & ROC AUC                 & 0.49  \\ \cline{2-3} 
 & Kappa Score             & 0.107 \\ \cline{2-3} 
 & Log Loss                & 21.61 \\ \hline
\end{tabular}
\label{Tab.2}
\end{table}

ResNet50:\\
Among four CNN model, ResNet50 performed best nevertheless not learning at the beginning. In the end, validation and training loss comes closer.\\
\begin{figure}[H]
  \centering
   
  \includegraphics[width=\linewidth]{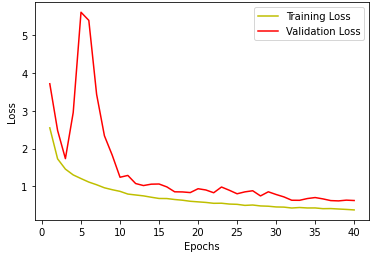}
 
  \Description{ResNet50
}
\caption{Cross Entropy loss of ResNet50
}
  \label{Fig.7}
\end{figure}

\begin{table}[H]
\caption{Score of ResNet50}
\begin{tabular}{|l|l|l|}
\hline
\multicolumn{1}{|c|}{Model} & \multicolumn{1}{c|}{Metric} & \multicolumn{1}{c|}{Score} \\ \hline
\multirow{7}{*}{ResNet50}   & F1 score (micro avg)        & 0.39                       \\ \cline{2-3} 
 & Hamming Score           & 0.387  \\ \cline{2-3} 
 & Hamming Loss            & 0.408  \\ \cline{2-3} 
 & Balanced Accuracy Score & 0.32  \\ \cline{2-3} 
 & ROC AUC                 & 0.49  \\ \cline{2-3} 
 & Kappa Score             & -0.014 \\ \cline{2-3} 
 & Log Loss                & 21.178 \\ \hline
\end{tabular}
\label{Tab.3}
\end{table}

InceptionResNetV2:\\
InceptionResNetV2 took a lot of time to complete training as the parameter is around 54 million. It was trained until epoch 30, but from 25 number epoch it started to fluctuate greatly.\\
\begin{figure}[h]
  \centering

  \includegraphics[width=\linewidth]{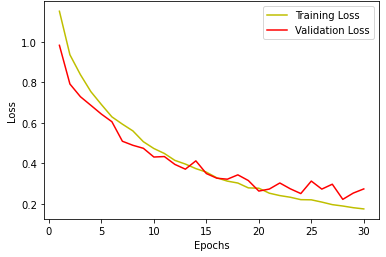}
 
  \Description{InceptionResNetV2}
     \caption{Cross Entropy loss of InceptionResNetV2}
  \label{Fig.8}
\end{figure}

\begin{table}[H]
\caption{Score of InceptionResNetV2}
\begin{tabular}{|l|l|l|}
\hline
\multicolumn{1}{|c|}{Model} & \multicolumn{1}{c|}{Metric} & \multicolumn{1}{c|}{Score} \\ \hline
\multirow{7}{*}{InceptionResNetV2}   & F1 score (micro avg)        & 0.37                       \\ \cline{2-3} 
 & Hamming Score           & 0.37  \\ \cline{2-3} 
 & Hamming Loss            & 0.42  \\ \cline{2-3} 
 & Balanced Accuracy Score & 0.31  \\ \cline{2-3} 
 & ROC AUC                 & 0.5  \\ \cline{2-3} 
 & Kappa Score             & -0.03 \\ \cline{2-3} 
 & Log Loss                & 21.7 \\ \hline
\end{tabular}
\label{Tab.4}
\end{table}

\textbf{Bagging algorithm:}\\
Random forest is a machine learning technique that combines many different decision trees to get a more accurate and reliable forecast. The ideal split for each node is determined using a set of randomly generated candidate variables during the tree-building process. Following trees in the bagging process are not dependent on prior trees, and each one is built separately using a bootstrap sample of the data set. Finally, a simple majority vote is used to make the prediction.

Random Forest Classifier was used with CNN models instead of the dense layer as the Neural Networks will require much more data. 200 trees were used in the forest before taking the maximum voting. The extracted features from CNN were flattened to feed into Random Forest Classifier along with the target variable.\\

\begin{table}[H]
\caption{Bagging Algorithm}
\label{Tab.5}
\resizebox{\columnwidth}{!}{%
\begin{tabular}{|c|cccc|}
\hline
                        & \multicolumn{4}{c|}{Model}                                                                \\ \hline
Metric &
  \multicolumn{1}{c|}{InceptionV3 + RandomForest} &
  \multicolumn{1}{c|}{ResNet50 + RandomForest} &
  \multicolumn{1}{c|}{Xception + RandomForest} &
  InceptionResNetV2 + RandomForest \\ \hline
F1 score (micro avg)    & \multicolumn{1}{c|}{0.85} & \multicolumn{1}{c|}{0.94}  & \multicolumn{1}{c|}{0.90} & 0.88 \\ \hline
Hamming Score           & \multicolumn{1}{c|}{0.84} & \multicolumn{1}{c|}{0.94}  & \multicolumn{1}{c|}{0.90} & 0.87 \\ \hline
Hamming Loss            & \multicolumn{1}{c|}{0.10} & \multicolumn{1}{c|}{0.03}  & \multicolumn{1}{c|}{0.06} & 0.08 \\ \hline
Balanced Accuracy Score & \multicolumn{1}{c|}{0.78} & \multicolumn{1}{c|}{0.92}  & \multicolumn{1}{c|}{0.86} & 0.82 \\ \hline
ROC AUC                 & \multicolumn{1}{c|}{0.85} & \multicolumn{1}{c|}{0.945} & \multicolumn{1}{c|}{0.90} & 0.87 \\ \hline
Kappa Score             & \multicolumn{1}{c|}{0.75} & \multicolumn{1}{c|}{0.9}   & \multicolumn{1}{c|}{0.84} & 0.80 \\ \hline
Log Loss                & \multicolumn{1}{c|}{5.38} & \multicolumn{1}{c|}{1.97}  & \multicolumn{1}{c|}{3.47} & 4.30 \\ \hline
\end{tabular}%
}
\end{table}

\textbf{Decision Tree:}
XGBoost is a well-known gradient boosting approach (ensemble) that improves the performance and speed of tree-based machine learning algorithms (sequential decision trees). In Ensemble Learning, XGBoost is classified as a boosting strategy. Ensemble learning combines different models into a collection of predictors to improve prediction accuracy. The faults created by prior models are attempted to be repaired by subsequent models by adding weights to the models in the boosting strategy.

XGBoost was used as the classifier along with CNN. 500 trees were used in the XGBoost. 

\begin{table}[H]
\caption{Decision Tree}
\label{Tab.6}
\resizebox{\columnwidth}{!}{%
\begin{tabular}{|c|cccc|}
\hline
                        & \multicolumn{4}{c|}{Model}                                                               \\ \hline
Metric & \multicolumn{1}{c|}{InceptionV3 + XgBoost} & \multicolumn{1}{c|}{ResNet50 + XgBoost} & \multicolumn{1}{c|}{Xception + XgBoost} & InceptionResNetV2 + XgBoost \\ \hline
F1 score (micro avg)    & \multicolumn{1}{c|}{0.91} & \multicolumn{1}{c|}{0.96} & \multicolumn{1}{c|}{0.93} & 0.92 \\ \hline
Hamming Score           & \multicolumn{1}{c|}{0.91} & \multicolumn{1}{c|}{0.95} & \multicolumn{1}{c|}{0.92} & 0.92 \\ \hline
Hamming Loss            & \multicolumn{1}{c|}{0.05} & \multicolumn{1}{c|}{0.02} & \multicolumn{1}{c|}{0.04} & 0.05 \\ \hline
Balanced Accuracy Score & \multicolumn{1}{c|}{0.88} & \multicolumn{1}{c|}{0.94} & \multicolumn{1}{c|}{0.90} & 0.90 \\ \hline
ROC AUC                 & \multicolumn{1}{c|}{0.91} & \multicolumn{1}{c|}{0.96} & \multicolumn{1}{c|}{0.93} & 0.93 \\ \hline
Kappa Score             & \multicolumn{1}{c|}{0.86} & \multicolumn{1}{c|}{0.93} & \multicolumn{1}{c|}{0.88} & 0.88 \\ \hline
Log Loss                & \multicolumn{1}{c|}{3.09} & \multicolumn{1}{c|}{1.47} & \multicolumn{1}{c|}{2.48} & 2.65 \\ \hline
\end{tabular}%
}
\end{table}

\textbf{Support Vector Machine}
The purpose of the Linear SVC (Support Vector Classifier) is to fit the data provided and generate a "best fit" hyperplane that divides or categorizes the data. Following that, we can input some features to the classifier to check what the "predicted" class is after we've obtained the hyperplane. Despite the fact that SVC and Linear SVC are supposed to optimize the same issue, all liblinear estimators penalize the intercept, but libsvm estimators do not. LinearSVC's underlying estimators are liblinear, which penalizes the intercept. SVC makes use of libsvm estimators, which do not. Liblinear estimators are tailored for a linear (special) scenario, hence they converge faster than libsvm on vast amounts of data. This is because the linear kernel is a particular situation that is optimized for in Liblinear but not in Libsvm. The One-vs-All (also known as One-vs-Rest) multiclass reduction is used by LinearSVC, whereas the One-vs-One multiclass reduction is used by SVC. SVC fits N * (N - 1) / 2 models to multi-class classification problems, where N is the number of classes. Linear SVC, on the other hand, only fits N models.

Linear SVC were employed as the classifier for the CNN’s extracted features.

\begin{table}[]
\caption{Support Vector Machine}
\label{Tab.7}
\resizebox{\columnwidth}{!}{%
\begin{tabular}{|c|cccc|}
\hline
                        & \multicolumn{4}{c|}{Model}                                                                 \\ \hline
Metric &
  \multicolumn{1}{c|}{InceptionV3 + Linear SVC} &
  \multicolumn{1}{c|}{ResNet50 + Linear SVC} &
  \multicolumn{1}{c|}{Xception + Linear SVC} &
  InceptionResNetV2 + Linear SVC \\ \hline
F1 score (micro avg)    & \multicolumn{1}{c|}{0.90}  & \multicolumn{1}{c|}{0.93}  & \multicolumn{1}{c|}{0.91} & 0.92 \\ \hline
Hamming Score           & \multicolumn{1}{c|}{0.897} & \multicolumn{1}{c|}{0.925} & \multicolumn{1}{c|}{0.92} & 0.92 \\ \hline
Hamming Loss            & \multicolumn{1}{c|}{0.07}  & \multicolumn{1}{c|}{0.04}  & \multicolumn{1}{c|}{0.05} & 0.05 \\ \hline
Balanced Accuracy Score & \multicolumn{1}{c|}{0.88}  & \multicolumn{1}{c|}{0.925} & \multicolumn{1}{c|}{0.91} & 0.90 \\ \hline
ROC AUC                 & \multicolumn{1}{c|}{0.91}  & \multicolumn{1}{c|}{0.945} & \multicolumn{1}{c|}{0.93} & 0.93 \\ \hline
Kappa Score             & \multicolumn{1}{c|}{0.84}  & \multicolumn{1}{c|}{0.88}  & \multicolumn{1}{c|}{0.88} & 0.87 \\ \hline
Log Loss                & \multicolumn{1}{c|}{3.55}  & \multicolumn{1}{c|}{2.56}  & \multicolumn{1}{c|}{2.62} & 2.78 \\ \hline
\end{tabular}%
}
\end{table}

\begin{figure*}[h]
  \centering
   
  \includegraphics[width=\linewidth]{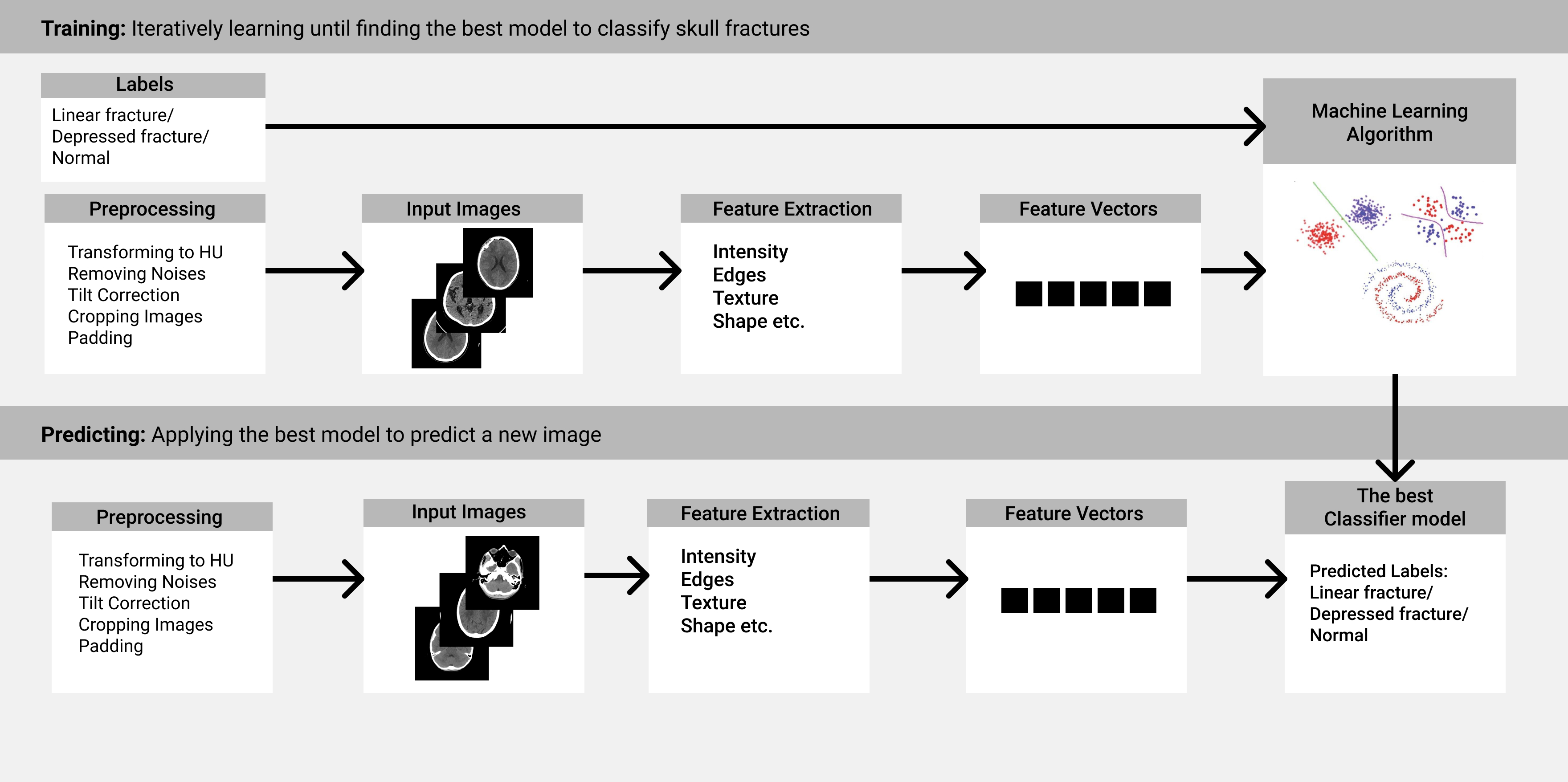}
 
  \Description{Decision Tree}
  \caption{Model Training Pipeline}
  \label{Tab.6}
\end{figure*}

\subsection{Metric}
Because there are so many classes to predict, the notion of positive and negative words and associated terms are calculated for each kind in a one vs. rest way, and then the overall levels are averaged. As a result, we chose F1 scores to evaluate all models.
   
 

  TP = True Positive, TN = True Negative, FP = False positive, FN = False Negative
 
\begin{equation}
{ Accuracy }=\frac{(T P+T N)}{(T P+F P+T N+F N)}
\end{equation}

\begin{equation}
{ Precision }=\frac{ TP}{( TP + FP)}
\end{equation}

\begin{equation}
{ Recall }=\frac{ TP }{( TP + FN )}
\end{equation}

    \begin{equation}
    F1=2 * \frac{{ Precision * Recall }}{{ Precision + Recall }}
    \end{equation}
\\
Balanced Accuracy, Hamming Loss, Hamming Score, ROC AUC, Kappa score, Log Loss were also used to check the model performance.
\subsection{Algorithmic details of the proposed system }
We developed a model that comprises ResNet50 and XGBoost. By combining ResNet50 as a pre-trained feature extractor to automatically obtain features from input and XGBoost as a classifier on the top level of the network to provide results, the ResNet50-XGBoost model gives more precise output. All of the ResNet50's feature extraction layers, as well as the feature flattening layer, are kept. The XGBoost model takes the role of the fully connected neural network and performs the classification task using the extracted features. ResNet50 was used to construct feature vectors from raw images, which gives a low-dimensional and noise-resistant manner to represent these images. Images were loaded into a pre-trained ResNet50 to build feature vectors, and the representation for that image in the intermediate layers of the neural network was utilized. At the penultimate layer of the ResNet50 model pre-trained on the ImageNet dataset, our approach extracts representations of given images. XgBoost was utilized in the classification phase to create direct predictions based on the high-level features extracted by ResNet50.

\section{Experimental Environments}
Ubuntu 20.04.2 LTS was utilized as the operating system. The AMD Ryzen Threadripper 1950X 16-Core Processor was used. The primary memory was 64 GB, and the graphics cards were two GeForce RTX 2080 Ti with 11 GB RAM each. TensorFlow 2.5.0rc2 was used as the deep learning framework.
\section{Experimental Result}
The dataset divided into three parts: 50\% training, 20\% validation and 30\% testing. It was made sure no slices from one patient was present in all three dataset. \hyperref[Tab.1]{Table 1}\hyperref[Tab.2]{,2} \hyperref[Tab.3]{,3}\hyperref[Tab.13]{,4} \hyperref[Tab.5]{,5} \hyperref[Tab.6]{,6} \hyperref[Tab.7]{,7} shows a comparison of the most well-known CNN models and CNN with supervised machine learning algorithm’s classification performance in terms of micro average F1 Score, Hamming Score, Hamming Loss, Balanced Accuracy, ROC AUC, Kappa score and Log Loss.

From these 7 tables, we observe that fine tuned ResNet50 combined with XGBoost achieves the highest score in all of the metrics. It is likely because every other CNN model stopped to learn better after 25 epoch, but ResNet50 was still learning quite well at 30th epoch.

ResNet50+XGBoost model’s classwise score is demonstrated in \hyperref[Tab.8]{Table 8}:

\begin{table}[H]
\caption{Classwise Score of Resnet50+XGBoost}
\label{Tab.8}
\begin{tabular}{|ccccc|}
\hline
\multicolumn{1}{|c|}{}                   & \multicolumn{1}{c|}{precision} & \multicolumn{1}{c|}{recall} & \multicolumn{1}{c|}{f1-score} & support \\ \hline
\multicolumn{1}{|c|}{Depressed Fracture} & \multicolumn{1}{c|}{0.93}      & \multicolumn{1}{c|}{0.98}   & \multicolumn{1}{c|}{0.98}     & 2548    \\ \hline
\multicolumn{1}{|c|}{Linear Fracture}    & \multicolumn{1}{c|}{0.96}      & \multicolumn{1}{c|}{0.87}   & \multicolumn{1}{c|}{0.91}     & 1373    \\ \hline
\multicolumn{1}{|c|}{Not Fractured} & \multicolumn{1}{c|}{0.99} & \multicolumn{1}{c|}{0.98} & \multicolumn{1}{c|}{0.99} & 2588 \\ \hline
\multicolumn{5}{|c|}{}                                                                                                         \\ \hline
\multicolumn{1}{|c|}{micro avg}     & \multicolumn{1}{c|}{0.96} & \multicolumn{1}{c|}{0.96} & \multicolumn{1}{c|}{0.96} & 6509 \\ \hline
\multicolumn{1}{|c|}{macro avg}     & \multicolumn{1}{c|}{0.96} & \multicolumn{1}{c|}{0.94} & \multicolumn{1}{c|}{0.95} & 6509 \\ \hline
\multicolumn{1}{|c|}{weighted avg}  & \multicolumn{1}{c|}{0.96} & \multicolumn{1}{c|}{0.96} & \multicolumn{1}{c|}{0.96} & 6509 \\ \hline
\multicolumn{1}{|c|}{samples avg}   & \multicolumn{1}{c|}{0.96} & \multicolumn{1}{c|}{0.96} & \multicolumn{1}{c|}{0.96} & 6509 \\ \hline
\end{tabular}
\end{table}

It is clear that our experimented model achieves a higher score in all three classes significantly.

\section{Discussion}
To our knowledge, this is the first research to disclose the development of a system that can distinguish between three types of skull fractures and test it with a small number of samples. To clarify this, \hyperref[tab:9]{Table 9} shows a comparison with existing published reference models vs our proposed work. Radiology reports were employed as ground labels to increase classification performance over CT scan images alone and supervised deep learning architectures for adding auxiliary data during training were presented. The suggested approach was successful in categorizing a CT scan of the head depending on the kind of skull fracture.  Despite having fewer image data than prior classification tests, our model performed wonderfully. Because the extracted features of images by CNN were fed into a powerful gradient boosting (GB) classifier that is an ensemble learning algorithm, which combines the predictions of multiple base learners (usually, each one being a fairly weak performer on its own) to generate one overall prediction for each input/example. This allows it to learn more complex relationships between the features and labels in the training set. This is a solid method when the dataset is very structured. As such, the final ensemble sequence can achieve (nearly) arbitrarily good performance on the training set. According to the outcomes of this investigation, our architecture has the potential to improve classification performance with high accuracy and F1 score. Because we have many more features than examples in our dataset, well-known CNN models with neural networks failed to perform effectively in this challenging multi-class classification task. We had a large number of weights to estimate, but the neural network couldn't since the NN's massive structure couldn't generalize effectively on our small, unbalanced medical dataset, and we required more data to learn such a large number of hidden weights. This is frequent in medical image processing since the design of well-known CNN models was too dense to learn from a small number of images, and they were not pre-trained on medical images. Despite having a higher overall F1-score, our dataset contains fewer Linear Fracture slices. As a result, our model did not perform as well on Linear Fracture as it did on other fractures, demonstrating that even with highly structured data, training with a short dataset by gradient boosters combined with CNN is difficult.
\begin{table}[H]
\begin{center}
\caption{Comparative performance of published model vs proposed model}

\setlength{\arrayrulewidth}{0.1mm}
 \setlength{\tabcolsep}{11pt}
\renewcommand{\arraystretch}{1.1}
\begin{tabular}{|p{1.5cm}|p{2.5cm}|p{2.2cm}|}
\hline
 Paper &Objective  &Score, Metric  \\
\hline
Shao and Zhao\cite{shao2003automatic}&Automatically detect if the skull is fractured or not  &100\%, accuracy  \\
\hline
Zaki et al.\cite{za20ki09new}&Segment fractured skull from 2D-CT brain image  &95\%, Normalized Euclidean Recall rate  \\
\hline

Yamada et al.\cite{yamada2016preliminary}&Detection of Linear skull fracture &80\% accuracy for a crack line of width 1.05mm  \\
\hline
Chilamkurthy et al. \cite{chilamkurthy2018deep}&Detection of multiple Hemorrhage and skull fracture of only calvaria. &91.11\%, AUC \\
\hline
Lee et al. \cite{lee2020classification}&Detection of femur fracture &86.78\%, accuracy  \\
\hline
Kuang et al. \cite{kuang2020skull}&Faster detection of skull fracture more accurately  &80\%, precision recall score  \\
\hline
Ours&Classification of three skull fractures &96\%, F1-score  \\

\hline

\end{tabular}

\end{center}

\label{tab:9}

\end{table} Medical image processing software has a lot of capabilities. Medical image analysis software can correctly detect inconsistencies and identify potentially harmful anomalies using Machine Learning methods. These diagnostic technologies can take over some of the most time-consuming processes, allowing clinicians to focus on issues that require immediate treatment. As a result, proper application of AI technology might assist healthcare organizations in providing better and more timely treatment.

\section{Conclusion}
Using head CT images, we developed a deep learning method to automatically identify and classify skull fractures. The proposed method for automatically classifying head CT images is very fast and human-level accurate. In contrast to previous research that focused just on detecting skull fractures, our goal was to categorize skull fractures from images into three different class. Our model classifies skull fracture images trained with limited data, so it has the advantages of effective utilization of little training data. We hope that by applying our technique to head CT images, we will be able to automate the head CT scan triage process. Our approach has the potential to make radiologists’ jobs easier. This technique has shown to have a promising future in delivering second opinions to doctors and radiologists. The ongoing enhancement of the algorithm is one of our major areas of focus for future research. To balance the multi-class dataset, other improvements could be done, such as the addition of GAN and SMOTE method.
\begin{acks}
	Syed Md. Shakawath Hossain, Medical Technologist, CT Scan department, Medinova Medical Services Ltd.; Md Sad Udiin Sadik, assistant manager; MD Ismail Hossain, Medical Technologist, CT Scan Department, Ibn Sina Hospital Sylhet Limited; Md Sad Udiin Sadik, assistant manager; MD Ismail Hossain, Medical Technologist, CT Scan department, Ibn They provided the CT brain images utilized in this study with the approval of the appropriate authorities. Dr. Tahmina Sumi of Z. H. Sikder Women's Medical College, Professor Dr. Ashikur Rahman Majumder, HEAD OF THE DEPARTMENT, Department of Radiology \& Imaging of Sylhet MAG Osmani Medical College, and Dr. Sajal Chandra Das of Ibn Sina Hospital Sylhet Limited, along with the respective radiologists, provided additional annotations of CT scan images. We'd also like to thank Sakib Alam Snigdha for letting us use his PC for our preliminary calculations.
\end{acks}

\bibliographystyle{ACM-Reference-Format}
\bibliography{sample-base}

\appendix

\end{document}